\documentclass[lettersize,journal]{IEEEtran}
\usepackage{amsmath,amsfonts}
\usepackage{algorithmic}
\usepackage{algorithm}
\usepackage{array}
\usepackage[caption=false,font=normalsize,labelfont=sf,textfont=sf]{subfig}
\usepackage{booktabs}
\usepackage{textcomp}
\usepackage{stfloats}
\usepackage{url}
\usepackage{verbatim}
\usepackage{graphicx}
\usepackage{cite}
\usepackage{amssymb}
\usepackage{multirow}
\usepackage{booktabs}
\usepackage{graphicx}
\usepackage{color,xcolor}
\usepackage{hyperref}
\usepackage{threeparttable}
\usepackage{multirow}
\usepackage{amsmath}
\makeatletter
\renewcommand{\maketag@@@}[1]{\hbox{\m@th\normalsize\normalfont#1}}%
\makeatother

\begin{document}

\title{WDMamba: When Wavelet Degradation Prior Meets Vision Mamba for Image Dehazing}

\author{%
  Jie Sun, %
  Heng Liu, %
  Yongzhen Wang, %
  Xiao-Ping Zhang, \textit{Fellow}, \textit{IEEE}, %
  and Mingqiang Wei, \textit{Senior Member}, \textit{IEEE}%
  
  \thanks{Jie Sun and Heng Liu contributed equally to this work.}%
  \thanks{Jie Sun, Heng Liu and Yongzhen Wang are with the School of Computer Science and Technology, Anhui University of Technology, Ma’anshan 243032, China (e-mail: jiesun@ahut.edu.cn; hengliu@ahut.edu.cn; wangyz@ahut.edu.cn).}%
  \thanks{Xiao-Ping Zhang is with the Tsinghua Shenzhen International Graduate School, Tsinghua University, Shenzhen 518055, China (e-mail: xpzhang@ieee.org).}%
  \thanks{Mingqiang Wei is with the School of Computer Science and Technology, Nanjing University of Aeronautics and Astronautics, Nanjing 210016, China, and also with the College of Artificial Intelligence, Taiyuan University of Technology, Taiyuan 030024, China (e-mail: mingqiang.wei@gmail.com).}%
}

\markboth{Journal of \LaTeX\ Class Files,~Vol.~14, No.~8, August~2021}%
{Shell \MakeLowercase{\textit{et al}}: A Sample Article Using IEEEtran.cls for IEEE Journals}


\maketitle

\begin{abstract}
In this paper, we reveal a novel haze-specific wavelet degradation prior observed through wavelet transform analysis, which shows that haze-related information predominantly resides in low-frequency components. Exploiting this insight,  we propose a novel dehazing framework, WDMamba, which decomposes the image dehazing task into two sequential stages: low-frequency restoration followed by detail enhancement. This coarse-to-fine strategy enables WDMamba to effectively capture features specific to each stage of the dehazing process, resulting in high-quality restored images. Specifically, in the low‐frequency restoration stage, we integrate Mamba blocks to reconstruct global structures with linear complexity,  efficiently removing overall haze and producing a coarse restored image. Thereafter, the detail enhancement stage reinstates fine‐grained information that may have been overlooked during the previous phase, culminating in the final dehazed output. Furthermore, to enhance detail retention and achieve more natural dehazing, we introduce a self-guided contrastive regularization during network training. By utilizing the coarse restored output as a hard negative example, our model learns more discriminative representations, substantially boosting the overall dehazing performance. Extensive evaluations on public dehazing benchmarks demonstrate that our method surpasses state-of-the-art approaches both qualitatively and quantitatively. Code is available at \textcolor{magenta}{\href{https://github.com/SunJ000/WDMamba}{https://github.com/SunJ000/WDMamba}}.

\end{abstract}


%-------------------------------------------------------------------------
\section{Introduction}
\IEEEPARstart{T}{he} presence of haze leads to substantial degradation of visual information, resulting in impaired image quality. This issue not only impairs human visual perception but also significantly compromises the performance of high-level vision tasks, including but not limited to object detection \cite{liu2018improved} and semantic segmentation \cite{ren2018deep}. Consequently, restoring sharp and clean images from haze-affected scenes is a critical task in computer vision research.

Conventional image dehazing methodologies typically rely on hand-crafted priors \cite{he2010single, meng2013efficient, zhu2015fast, berman2016non, ju2021idbp}, derived from statistical observations, to estimate the transmission map or global atmospheric light. These methods then restore haze-free images based on the atmospheric scattering model \cite{narasimhan2000chromatic}. However, these priors often fail to generalize across all scenarios, resulting in common issues such as incomplete dehazing and color distortion.

With the advent of large-scale synthetic datasets and advancements in deep learning techniques, mainstream image dehazing methods have transitioned to design sophisticated Convolutional Neural Networks (CNNs) or Transformer networks that directly learn the mapping from hazy to haze-free images in an end-to-end manner. Despite these significant advances, several challenges persist that limit their performance. First, CNN-based methods are inherently limited by their constrained receptive fields, rendering them less effective in capturing global dependencies within the data. Transformer-based approaches, although leveraging the multi-head self-attention mechanism to attain a global receptive field, unavoidably introduce significant computational overhead, demanding substantial computational resources. Hence, the adoption of more efficient and advanced techniques is essential to facilitate image dehazing. Recently, the vision state space model Mamba \cite{liu2024vmamba} has garnered considerable interest owing to its capacity to capture long-range dependencies with linear complexity, yielding compelling performance in domains such as medical image segmentation \cite{ma2024u} and image deraining \cite{zou2024freqmamba}. However, within the domain of image dehazing, the investigation of this technique remain nascent.
\begin{figure}[t] 
    \centering
    \includegraphics[width=1.0\linewidth]{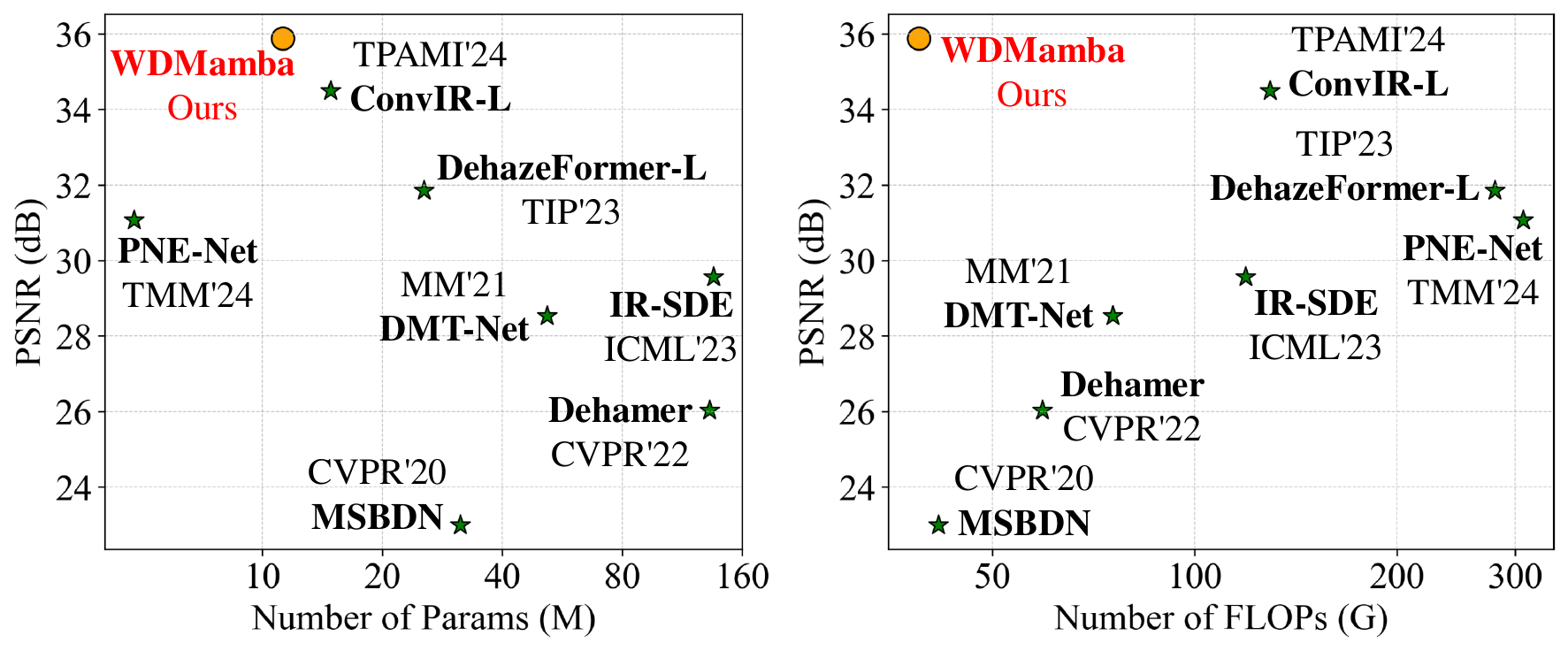}
\caption{Comparison of model performance and complexity on the Haze4K dataset \cite{liu2021synthetic} across various state-of-the-art methods. FLOPs are calculated at a resolution of 256\( \times \)256.}
 	\label{fig:param}
\end{figure}

\begin{figure}[t] 
    \centering
    \includegraphics[width=1.0\linewidth]{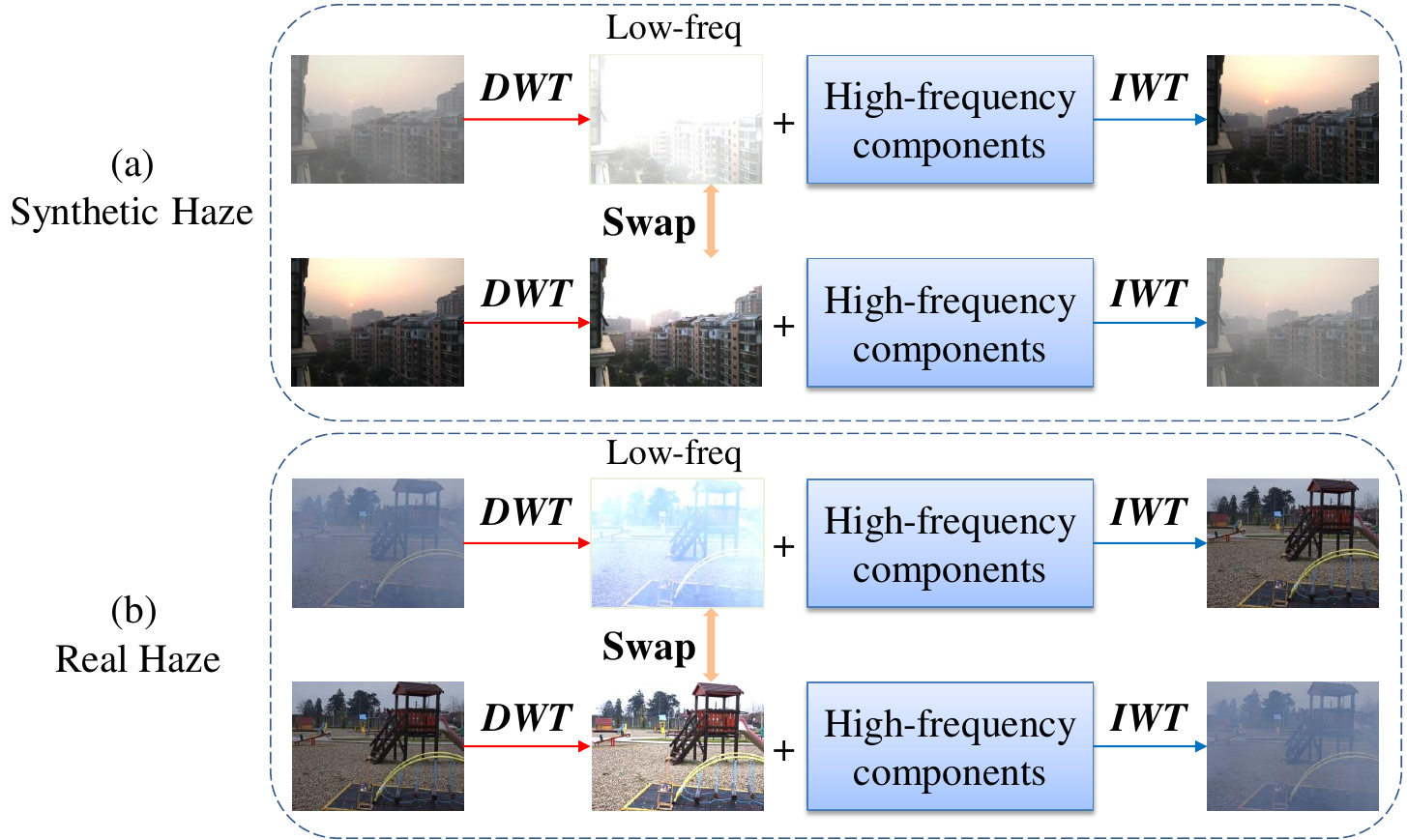}
	\caption{Wavelet degradation prior of haze. By exchanging the low-frequency subbands between paired hazy and haze-free images during wavelet decomposition, it can be observed that haze-related degradation is predominantly concentrated in the low-frequency components. DWT and IWT denote the Discrete Wavelet Transform and the Inverse Wavelet Transform, respectively, while Low-freq refers to the low-frequency subband of the decomposed image. 
}
 	\label{fig:WDPrior}
\end{figure}

Despite notable advancements in learning-based approaches, the intrinsically ill-posed nature of image dehazing endures as a fundamental challenge that existing techniques fail to address satisfactorily. Investigating the inherent physical properties of haze degradation can empower networks to more effectively capture features essential for robust dehazing. Nevertheless, this critical perspective has been largely neglected by prevailing methods, thereby heightening the risk of overfitting. To mitigate this issue, we observe a haze-specific wavelet degradation prior by analyzing the wavelet decomposition and reconstruction process of paired hazy and clean images. As depicted in Fig. \ref{fig:WDPrior}, the degradation is markedly suppressed when reconstructing the image using the low-frequency components of the clean image and the high-frequency components of the hazy image, suggesting that haze-induced degradation predominantly resides in the low-frequency components. This insight inspires a novel formulation of the dehazing problem as two decoupled sub-tasks: low-frequency restoration and high-frequency detail enhancement. Such a decomposition enables the network to more effectively learn task-adaptive feature representations at each stage, thereby facilitating the generation of high-fidelity dehazed images.

In this paper, we propose a novel wavelet degradation prior-guided Mamba framework for efficient image dehazing, referred to as WDMamba. Leveraging this prior knowledge, WDMamba is capable of learning task-adaptive feature representations more effectively during both the low-frequency restoration and subsequent detail enhancement stages, thereby achieving superior restoration performance. Concretely, given a hazy image, a Mamba-based Low-Frequency Restoration Network (LFRN) is first employed to operate independently on the degraded low-frequency components, efficiently reconstructing global structures with linear complexity to yield a coarsely restored image. This is followed by a CNN-based Detail Enhancement Network (DEN), which refines the output by recovering high-frequency details that are inadequately restored in the preceding stage, ultimately producing the final dehazed result. Furthermore, to enhance the perceptual realism of the dehazed outputs, we introduce a Self-Guided Contrastive Regularization (SGCR) paradigm, wherein the coarse restorations are treated as hard negative samples, enabling the network to mine informative cues from itself and promote the generation of more natural and visually plausible images. Fig. \ref{fig:param} presents a comparative analysis between our WDMamba and state-of-the-art approaches in terms of both performance and computational complexity. The main contributions of this work are outlined as:
\begin{itemize}
    \item We propose WDMamba, a novel Wavelet Degradation prior-guided Mamba for efficient image dehazing. By exploiting the inherent physical properties of the degradation process, it attains superior dehazing performance through the decomposition of the task into two complementary stages, enabling more effective learning of task-adaptive feature representations across distinct phases.
    \item We present a Low-Frequency Restoration Network (LFRN) that specifically designed to address the low-frequency components. By incorporating Mamba blocks, LFRN enables effective global structure restoration with linear computational complexity, thus striking an optimal balance between performance and efficiency.
    \item We introduce a novel self-guided contrastive regularization paradigm, wherein the coarsely restored image serves as effective guidance for accurately recovering fine-grained details.
\end{itemize}

\section{Related Work}
In this section, we provide a review of image dehazing methods, state space models, and contrastive learning techniques in image restoration.
\subsection{Single Image Dehazing}
\subsubsection{Prior-Based Methods} Traditional image dehazing methods relied on prior knowledge derived from statistical observations, and subsequently recovered haze-free images based on the atmospheric scattering model \cite{narasimhan2000chromatic}. For instance, He et al. \cite{he2010single} observed that in the non-sky regions of haze-free images, certain pixels consistently exhibit low values in at least one color channel, leading to the proposal of the dark channel prior (DCP) for image dehazing. Additionally, Zhu et al. \cite{zhu2015fast} observed that haze density is positively correlated with the difference in brightness and saturation of the scene, and proposed a color attenuation prior (CAP). Berman et al. \cite{berman2016non}, based on the assumption that the colors of haze-free images are well approximated by a few hundred distinct colors, which form tight clusters in the RGB space, introduced a non-local image dehazing method. Although these methods have improved the overall visual quality of hazy images, these priors do not effectively capture the fine-grained details within the images, making it challenging to generate crisp outputs.

\subsubsection{Learning-Based Methods} Benefiting from the significant breakthroughs of deep learning technologies and the availability of large-scale synthetic datasets, mainstream image dehazing methods have shifted towards directly recovering haze-free images in an end-to-end manner by designing sophisticated CNN or Transformer networks. Liu et al. \cite{liu2019griddehazenet} proposed a multi-scale dehazing network based on the attention mechanism, enabling efficient interaction among information at different scales. Dong et al. \cite{dong2020multi} introduced the back-projection technique from image super-resolution into the dehazing task and proposed a multi-scale enhancement dehazing network with dense feature fusion based on the U-Net architecture. Qin et al. \cite{qin2020ffa} combined channel attention with pixel attention mechanisms to construct a feature attention block, and proposed a feature fusion attention network. Zhang et al. \cite{zhang2021hierarchical} integrated hierarchical feature fusion with hybrid convolutional attention to progressively enhance dehazing performance. Compared to purely CNN-based image dehazing methods, Zhang et al. \cite{zhang2023sdbad} proposed a spatial dual-branch attention dehazing network based on the Meta-Former paradigm. Song et al. \cite{song2023vision}, building upon Swin Transformer \cite{liu2021swin}, introduced DehazeFormer, addressing the limitations of the original design for dehazing tasks. Wang et al. \cite{wang2023uscformer} designed a unified Transformer-CNN architecture that simultaneously captures global and local feature dependencies to enhance dehazing performance.

\subsection{State Space Models}
State Space Models (SSMs) \cite{gu2021efficiently, smith2022simplified}, which originate from systems control theory, have attracted increasing attention for their strong ability to model long-range dependencies while preserving linear scalability. Recently, the emergence of Mamba \cite{gu2023mamba}, a novel SSM that integrates hardware-aware algorithms with selective state update mechanisms, has demonstrated superior performance and greater computational efficiency compared to Transformer models in the field of natural language processing. Moreover, Mamba has exhibited promising adaptability within the field of computer vision, with successful applications in biomedical image segmentation \cite{ma2024u}, image deraining \cite{zou2024freqmamba}, and low-light image enhancement \cite{zou2024wave}. In this study, we take the initiative to explore the application of Mamba to the task of image dehazing.

\subsection{ Contrastive Learning for Image Restoration}
Contrastive learning \cite{oord2018representation,sermanet2018time,chen2020simple,he2020momentum} has achieved significant advancements in self-supervised and unsupervised learning tasks by evaluating the similarity and divergence between different samples in the feature space. Additionally, the integration of contrastive regularization terms has been demonstrated to enhance model performance in low-level vision tasks, such as image dehazing and image deraining. For instance, Zheng et al. \cite{zheng2023curricular} constructed negative samples of varying difficulty using an additional backbone model and proposed a curricular contrastive regularization, which effectively improves dehazing performance. Similarly, Feng et al. \cite{feng2024advancing} introduced a Gaussian perceptual contrastive loss to further constrain the network's updates towards the natural dehazing direction. Gao et al. \cite{gao2024efficient} proposed a frequency-domain contrastive regularization, which enhances the model's performance in image deraining tasks. Furthermore, Wang et al. \cite{wang2024ucl} developed a pixel-wise contrastive perceptual loss for unsupervised image dehazing. In this work, we introduce a self-guided contrastive regularization that leverages the coarse restored image as guidance, effectively enhancing the dehazing performance.
\begin{figure*}[t] 
    \centering
    \includegraphics[width=1.0\linewidth]{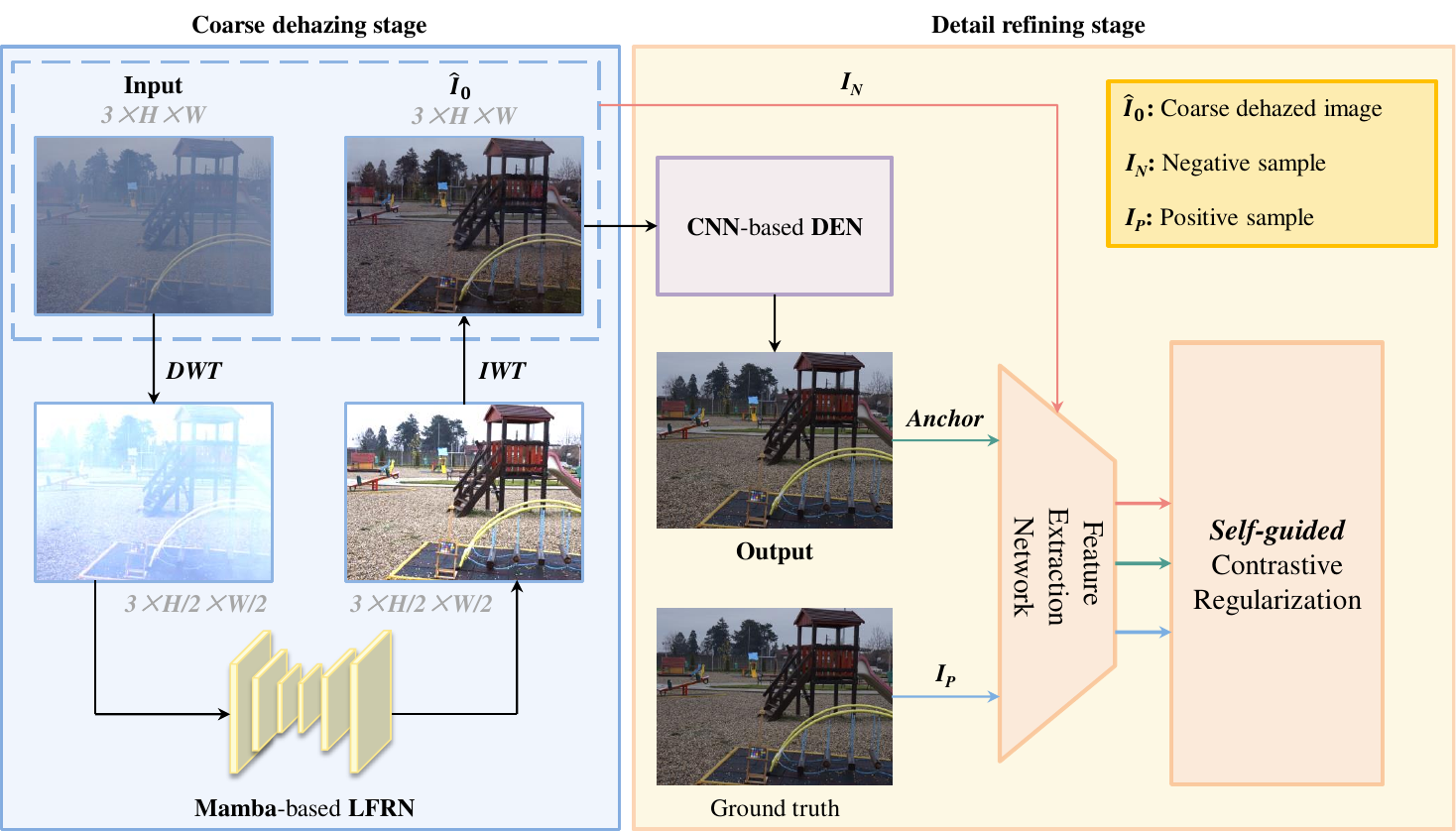}
	\caption{
	The overall framework of the proposed WDMamba. It consists of three main components: the Low-Frequency Restoration Network (LFRN), the Detail Enhancement Network (DEN), and the Self-Guided Contrastive Regularization (SGCR). LFRN operates on the degraded low-frequency components, employing linear complexity to restore global structures and produce a coarse restored image. Subsequently, DEN enhances local details, generating the final output. SGCR treats the coarse restored image as a hard negative sample, encouraging the final dehazed image to better approximate the clean image.
 	}
 	\label{fig:WDMamba}
 \end{figure*}
\section{METHOD}
Distinct from the majority of existing dehazing approaches, WDMamba integrates the intrinsic prior of haze degradation while leveraging the respective advantages of Mamba and CNN architectures. This design empowers the model to effectively capture both global contextual dependencies and fine-grained local details, thereby facilitating the generation of high-fidelity dehazed images. Moreover, a self-guided contrastive regularization is incorporated during training, wherein the coarsely restored outputs serve as guidance to iteratively refine the model's dehazing capability. In this section, we first present the necessary preliminaries, followed by an in-depth exposition of the algorithmic pipeline and the architecture details of the proposed framework.

\subsection{Preliminaries}
\textbf{State Space Models.} 
State Space Models (SSMs) represent a fundamental framework in control theory for characterizing the evolution of dynamic systems over time. They employ state variables to relate the inputs, outputs, and internal states of a system. Mathematically, such systems can be expressed using linear ordinary differential equations as: 
\begin{equation}
h'(t) = \mathbf{A} h(t) + \mathbf{B} x(t)
\end{equation}
\begin{equation}
y(t) = \mathbf{C} h(t) + \mathbf{D} x(t)
\end{equation}
where \( x(t) \) and \( y(t) \) represent the system’s input and output, respectively, \( h(t) \) denotes the state variable that characterizes the current state of the system. The matrices \( \mathbf{A} \), \( \mathbf{B} \), \( \mathbf{C} \), and \( \mathbf{D} \) are model parameters that define the system’s dynamic properties and the relationship between the input and output.

To integrate SSMs into deep learning algorithms, researchers employed the zero-order hold (ZOH) technique, aligning the model with the sampling rate of the underlying signals present in the input data, thereby enabling the transformation of discretized data into a continuous form. This process can be defined as:
\begin{equation}
h'_t = \bar{\mathbf{A}} h_{t-1} + \bar{\mathbf{B}} x_t
\end{equation}
\begin{equation}
y_t = \mathbf{C} h_t + \mathbf{D} x_t
\end{equation}
\begin{equation}
\bar{\mathbf{A}} = e^{\Delta \mathbf{A}}
\end{equation}
\begin{equation}
\bar{\mathbf{B}} = (\Delta \mathbf{A})^{-1} \left(e^{\Delta \mathbf{A}} - I\right) \cdot \Delta \mathbf{B}
\end{equation}
where \( \Delta \) represents a learnable time-scale parameter,and the matrices \( \bar{\mathbf{A}} \) and \( \bar{\mathbf{B}} \) correspond to the discrete-form parameters of \( \mathbf{A} \) and \( \mathbf{B} \), respectively. More recently, Mamba was proposed, featuring a simple yet effective selection mechanism that reparameterizes the SSM based on the input. This enables the model to filter out irrelevant information while preserving essential and relevant data. Furthermore, Mamba employs a parallel scanning algorithm to iteratively compute the model, ensuring high efficiency during both training and inference.

\textbf{Discrete Wavelet Transform.}
An RGB image \( I \in \mathbb{R}^{H \times W \times C} \) can be decomposed into four frequency subbands using the Haar wavelet transform, formally defined as:
\begin{equation}
\{cA, cH, cV, cD\} = DWT(I)
\end{equation}
Here, \( cA \in \mathbb{R}^{\frac{H}{2} \times \frac{W}{2} \times C} \) denotes the low-frequency component capturing global information, while \( cH \), \( cV \), and \( cD \) \( \in \mathbb{R}^{\frac{H}{2} \times \frac{W}{2} \times C} \) are high-frequency components encoding texture details in horizontal, vertical, and diagonal directions, respectively. These subbands can be reconstructed into the original image via the inverse wavelet transform, i.e.,
\begin{equation}
I = IWT(cA, cH, cV, cD)
\end{equation}
It is worth noting that no information is lost during the wavelet decomposition and reconstruction process.

\textbf{Discrete Fourier Transform.}
The Discrete Fourier Transform (DFT) is a widely utilized technique in image processing, enabling the conversion of discrete spatial-domain signals into their corresponding discrete frequency-domain representations, and vice versa using the inverse Discrete Fourier Transform (iDFT). For a 2D single-channel image \( x \) of size \( H \times W \), the DFT is defined as:
\begin{equation}
\mathcal{F}(x)(u,v) = \sum_{h=0}^{H-1} \sum_{w=0}^{W-1} x(h,w) e^{-j 2\pi \left( \frac{h}{H}u + \frac{w}{W}v \right)}
\end{equation}
where \( \mathcal{F}(x)(u,v) \) represents the frequency-domain representation of \( x(h,w) \), and \( u \) and \( v \) are the frequency indices. The result in the frequency domain consists of a real part \( \mathcal{R}(x) \) and an imaginary part \( \mathcal{I}(x) \), which can be used to compute the amplitude spectrum and phase spectrum as:
\begin{equation}
\mathcal{A}(x)(u,v) = \sqrt{\mathcal{R}^2(x)(u,v) + \mathcal{I}^2(x)(u,v)}
\end{equation}
\begin{equation}
\mathcal{P}(x)(u,v) = \arctan \left[ \frac{\mathcal{I}(x)(u,v)}{\mathcal{R}(x)(u,v)} \right]
\end{equation}
These two components collectively encode the intrinsic properties of the image, with the amplitude representing the global structure and the phase capturing fine-grained details. 
\subsection{Overview}
The overall architecture of our WDMamba is depicted in Fig. \ref{fig:WDMamba}, comprising a Low-Frequency Restoration Network and a Detail Enhancement Network. Given a hazy input image, the low-frequency components are initially extracted via discrete wavelet transformation and processed by a U-Net \cite{ronneberger2015u}-based Low-Frequency Restoration Network, which employs the Mamba mechanism to efficiently reconstruct global structures with linear computational complexity. An inverse wavelet transformation is then applied to yield a coarsely restored image. This intermediate result is subsequently refined by a CNN-based Detail Enhancement Network, which leverages the local feature extraction capability of CNNs and incorporates frequency-domain enhancement to further improve fine details. Additionally, we introduce a self-guided contrastive regularization mechanism, wherein the coarsely restored images serve as negative samples within a contrastive learning framework. This strategy significantly boosts the model's dehazing effectiveness, enabling the generation of visually natural and artifact-free outputs. In the subsequent section, we will provide a comprehensive breakdown of the core architectural components.

\begin{figure*}[t] 
    \centering
    \includegraphics[width=1.0\linewidth]{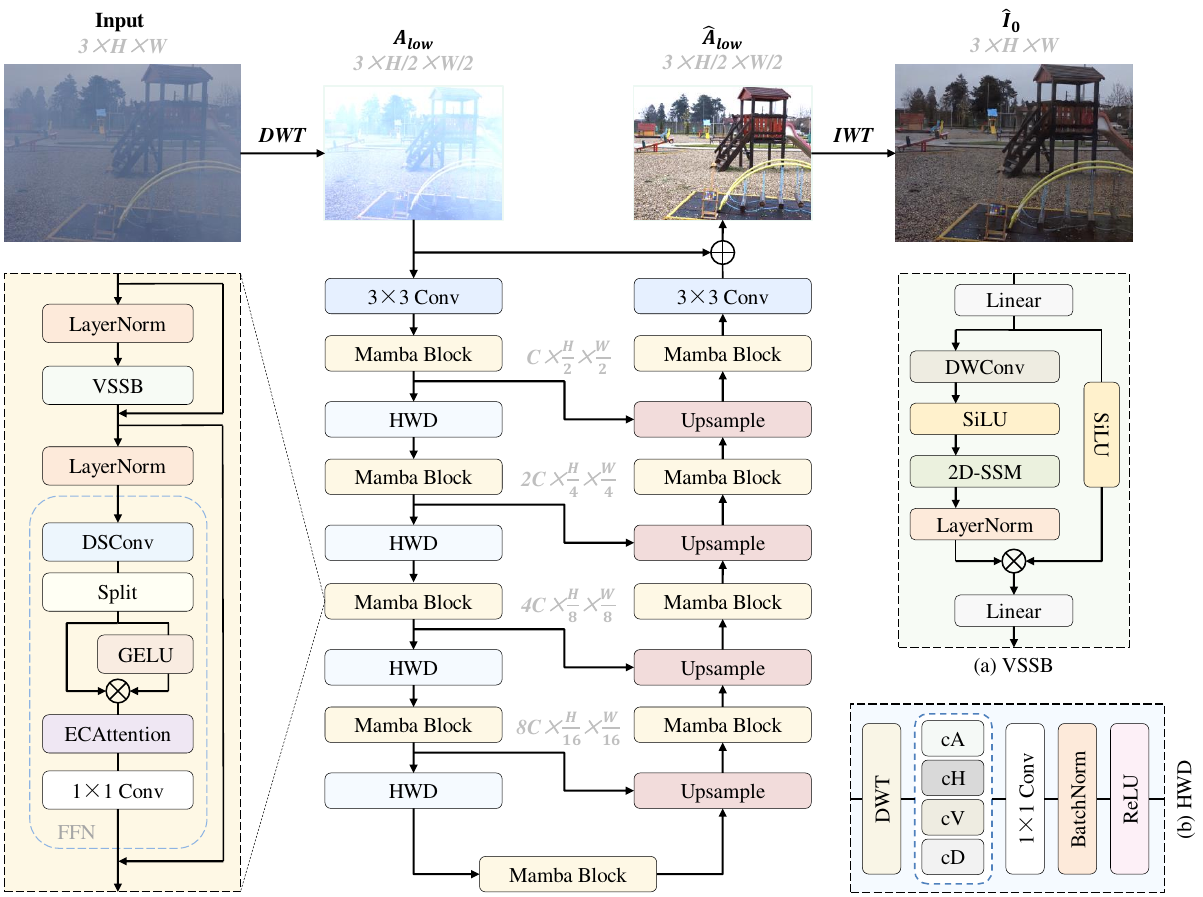}
	\caption{
	The detailed architecture of the low-frequency restoration network. It follows a U-Net design, incorporating Mamba blocks to model long-range feature dependencies for effective global information restoration. To further preserve structural integrity, Haar wavelet downsampling is employed in place of conventional downsampling operations, thereby alleviating information loss and enhancing feature fidelity throughout the restoration process.
 	}
 	\label{fig:LFRN}
 \end{figure*}
\subsection{Low-frequency Restoration Network} The detailed architecture of the low-frequency restoration network is illustrated in Fig. \ref{fig:LFRN}. It adopts a U-Net-like design, where Mamba blocks are integrated to leverage the strength of state space models in modeling long-range feature dependencies with linear complexity, thereby enabling efficient global structure restoration. In addition, we employ Haar Wavelet Downsampling (HWD) \cite{xu2023haar} to mitigate the spatial information loss typically introduced by conventional downsampling operations such as strided convolutions, thus facilitating more effective information transmission.

\noindent\( \mathbf{Mamba\ Block.} \) As illustrated in Fig. \ref{fig:LFRN}, the Mamba block primarily consists of a Visual State Space Block (VSSB) and a Feed-Forward Network (FFN). For the input feature \( F_A^{in} \), we first apply layer normalization (LN) to standardize the feature representation. The VSSB is then employed to capture global dependency features. Subsequently, the feature passes through an FFN to enable efficient information propagation, assisting the subsequent layers in the network hierarchy to focus on finer image details. Similarly, another layer normalization is applied before feeding the features into the FFN. The entire process can be formulated as:
\begin{equation}
F_h = VSSB\left(LN\left(F_A^{in}\right)\right) + \alpha \cdot F_A^{in}
\end{equation}
\begin{equation}
F_A^{out} = FFN\left(LN\left(F_h\right)\right) + \beta \cdot F_h
\end{equation}
where \( F_h \) denotes the intermediate hidden feature, \( LN(\cdot) \) represents the layer normalization operation, \( VSSB(\cdot) \) and \( FFN(\cdot) \) correspond to the VSSB and FFN mappings, respectively. \( \alpha \) and \( \beta \) are learnable skip-scaling factors.

\noindent\( \mathbf{Visual\ State\ Space\ Block.} \) The visual state space block offers the advantage of global modeling with linear computational complexity and has achieved notable success in various low-level vision tasks, including image deraining and low-light image enhancement. In this work, the structure of VSSB remains consistent with its foundational implementation in Vmamba \cite{liu2024vmamba}. As illustrated in Fig. \ref{fig:LFRN} (a), VSSB comprises two parallel branches. Given the input feature \( X_{in} \in \mathbb{R}^{H \times W \times C} \), a shared-weight linear layer is first applied to expand the feature channels to \( \gamma C \) respectively, where \( \gamma \) is a pre-initialized channel expansion factor. In the first branch, the features are directly passed through a SiLU activation function to preserve the original information. In the other branch, a depthwise convolution is initially applied to extract spatial level features, followed by SiLU activation to enhance the nonlinear representational capacity. Subsequently, a 2D selective scanning module (2D-SSM) is employed to model long-range spatial dependencies, followed by a layer normalization operation. The outputs of both branches are then fused via a Hadamard product. Finally, another linear layer is used to reduce the feature channels back to \( C \), producing an output with the same dimensionality as the original input. 
\begin{figure}[t] 
    \centering
    \includegraphics[width=1.0\linewidth]{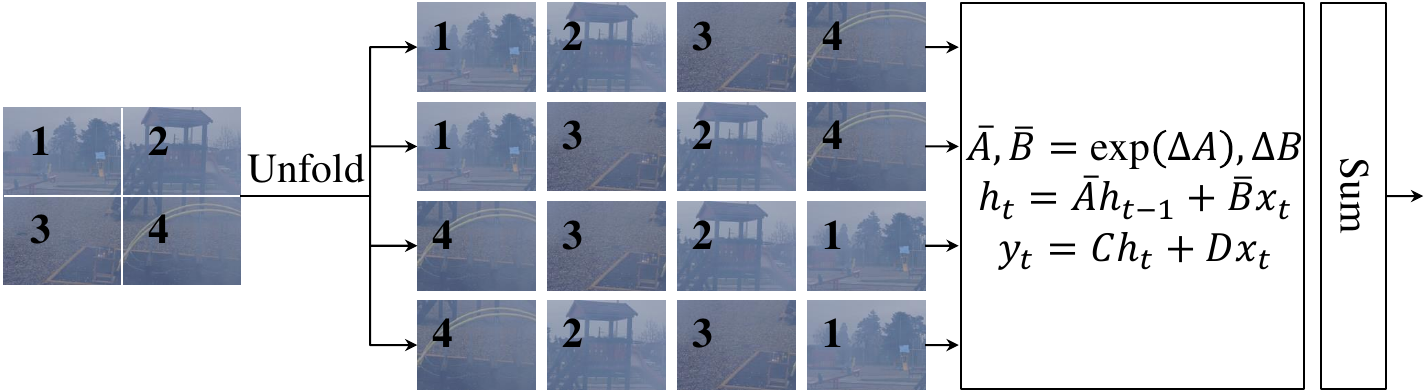}
	\caption{
	Illustration of 2D selective scanning mechanism.
 	}
 	\label{fig:SS2D}
 \end{figure}
 
\noindent\( \mathbf{2D\ Selective\ Scan\ (SS2D).} \) SS2D serves as a key mechanism that enables the VSSB effectively models long-range spatial dependencies. As illustrated in Fig. \ref{fig:SS2D}, given the input patches, four distinct scanning paths are employed to transform the image features into linear sequences, each of which is independently modeled by a discrete state space equation. Finally, the four sequences are aggregated through summation and reshaped back into 2D image features.

\noindent\( \mathbf{ Feed\text{-}Forward\ Network.}\) We employ a gated feed-forward network enhanced with an attention mechanism to regulate the flow of information within the Mamba block, assisting the subsequent network in focusing on more intricate details. The structure of the FFN is illustrated on the left side of Fig. \ref{fig:LFRN}. 
Given an input \( F_{in} \in \mathbb{R}^{H \times W \times C} \), a depthwise separable convolution is first employed to efficiently extract features across both channel and spatial dimensions. This is followed by a nonlinear gating mechanism that modulates the strength of information flow, facilitating the propagation of informative features to deeper layers while suppressing less relevant ones. Subsequently, an efficient channel attention (ECA) \cite{wang2020eca} module is applied to guide the network in focusing on the most relevant feature representations. Finally, a \( 1 \times 1 \) convolution is used to reduce the channel dimension back to \( C \). The entire process can be defined as:
\begin{equation}
F_{out} = {PConv}({ECA} (\delta_{G} ( {DSConv}(F_{in}))))
\end{equation}
where \( {PConv}(\cdot) \) and \( {DSConv}(\cdot) \) denote \( 1 \times 1 \) convolution and depthwise separable convolution, respectively; \( \delta_{G}(\cdot) \) is the nonlinear gating mechanism, and \( F_{out} \) represents the output.

\noindent\( \mathbf{ Haar\ Wavelet\ Downsampling.}\) In image restoration tasks, downsampling is typically conducted using strided convolution operations to aggregate local features and increase the receptive field. However, this may lead to the loss of important spatial information. Recently, Haar Wavelet Downsampling (HWD) has emerged as a novel alternative, demonstrating improved performance in semantic segmentation tasks. Inspired by this, we incorporate HWD into the low-frequency restoration network to better preserve essential information. The structure of HWD is illustrated in Fig. \ref{fig:LFRN} (b). First, the input features are transformed using the Haar wavelet transform, producing four subbands, each with half the spatial resolution of the original input. These subbands are then concatenated along the channel dimension. A \( 1 \times 1 \) convolution is subsequently applied to promote cross-channel information integration and interaction, followed by batch normalization and ReLU activation. By leveraging the spatial resolution reduction and the information-preserving property of the Haar wavelet transform, HWD enables downsampling without information loss, thereby enhancing the model’s dehazing performance.
\begin{figure*}[t] 
    \centering
    \includegraphics[width=1.0\linewidth]{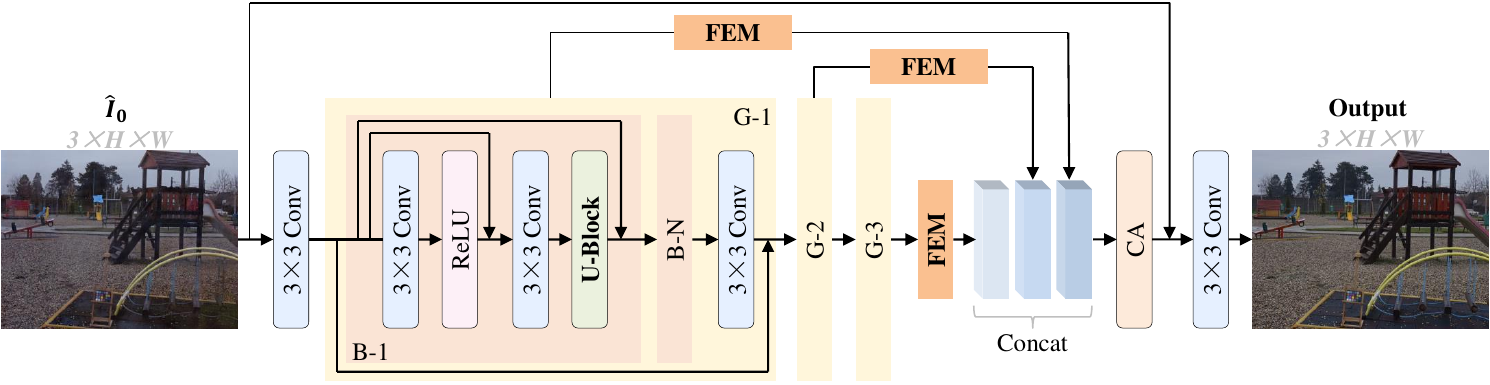}
	\caption{
	The detailed architecture of the detail enhancement network. It consists of three structural groups, each comprising \(N\) block structures. The block structure incorporates a miniature U-Net-style block (U-Block) to facilitate multi-scale feature fusion and enhance feature representation. Furthermore, the inclusion of the Frequency domain Enhancement Module (FEM) aids in recovering fine details by enhancing key frequency components.
 	}
 	\label{fig:DEN}
 \end{figure*}
\subsection{Detail Enhancement Network} The low-frequency restoration network is designed to recover only the degraded low-frequency components, resulting in a coarse restored image. To align fine-grained details, we propose a CNN-based Detail Enhancement Network (DEN) that further refines the coarse output, ultimately producing a sharp and clear dehazed image. Inspired by FFA-Net \cite{qin2020ffa}, DEN adopts a group-and-block architectural design. In addition, it explores frequency-domain information to facilitate a more accurate recovery of image details. As shown in Fig. \ref{fig:DEN}, for the coarse restored image, shallow feature extraction is initially conducted using the \( 3 \times 3 \) convolution, which also serves to expand the channel dimension. Subsequently, deeper feature extraction and enhancement are performed through three cascaded group structures, each comprising N feature enhancement blocks. The resulting features are further refined by Frequency Domain Enhancement modules (FEM). The hierarchical features output from the FEMs are then concatenated along the channel dimension, followed by the channel attention (CA) to adaptively adjust the feature representations. Finally, \( 3 \times 3 \) convolution is employed to map the features back to the RGB space, yielding the final dehazed image.

\noindent\( \mathbf{Block\ Structure.}\) The basic block structure is designed based on residual connections \cite{he2016deep} and a U-Net \cite{ronneberger2015u} style block (Ublock). For the input feature \( Y_{in} \), it is first passed through a convolutional layer, followed by ReLU activation, introducing the first residual connection. The feature is then sequentially processed through another convolutional layer and the Ublock, with the second residual connection introduced. The operation of the block structure can be defined as:
\begin{equation}
F_1 = ReLU\left(Conv\left(Y_{in}\right)\right) + Y_{in}
\end{equation}
\begin{equation}
Y_{out} = Ublock\left(Conv\left(F_1\right)\right) + Y_{in}
\end{equation}
where \( {Conv}(\cdot) \) and \( {Ublock}(\cdot) \) represent operations performed by a \( 3 \times 3 \) convolution and the Ublock, respectively. \( F_1 \) denotes the intermediate feature after the first residual connection is introduced, and \( Y_{{out}} \) represents the output of the basic block structure. In our design, the incorporation of residual connections helps retain the original input information while integrating it with the transformed features learned by the network. The Ublock further refines the features through skip connections and multiscale processing, enabling the network to effectively capture both low-level visual features and high-level semantic representations, thereby enhancing the network’s feature representation capability.
\begin{figure}[t] 
    \centering
    \includegraphics[width=0.8\linewidth]{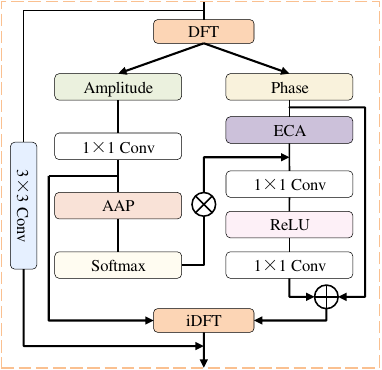}
	\caption{
	Specific implementation of frequency domain enhancement module.
 	}
 	\label{fig:FEM}
 \end{figure}
 
\noindent\( \mathbf{Frequency\ Domain\ Enhancement.}\) In the frequency domain of an image, especially in the phase components, a significant amount of detailed information is embedded. To better refine the coarse restored image, we introduce frequency domain enhancement in the detail enhancement network. The detailed structure of the frequency domain enhancement module is illustrated in Fig. \ref{fig:FEM}. Given the feature \( F_g \) obtained from the group structure, we first apply the discrete Fourier transform (DFT) to extract the amplitude spectrum \( \mathcal{A} \) and the phase spectrum \( \mathcal{P} \). The amplitude spectrum is then enhanced through \( 1\times1 \) convolution, expressed as \( \mathcal{A}' = PConv(\mathcal{A}) \). For the phase spectrum, the ECA mechanism is utilized to adaptively emphasize the importance of feature channels, after which the features are guided by the amplitude information and further refined through convolutional operations, formulated as:
\begin{equation}
\begin{aligned}
    &\omega_{\mathcal{A'}} = Softmax(AAP \left( \mathcal{A'} \right) ) \\
    &\mathcal{P}' = PConv(ReLU(PConv(( \omega_{\mathcal{A'}} \cdot ECA(\mathcal{P}))))) + \mathcal{P}
\end{aligned}
\end{equation}
where \( AAP(\cdot) \) denotes adaptive average pooling, \(\omega_{\mathcal{A'}}\) represents the attention weights of the enhanced amplitude spectrum. Finally, the frequency domain features are transformed back into the spatial domain via the inverse Fourier transform.

\subsection{Self-Guided Contrastive Regularization} In image dehazing, contrastive regularization treats the output of the dehazing network as the anchor, with the clean image \(J\) (ground truth) and hazy image \(I\) serving as positive and negative samples, respectively. The objective of contrastive regularization \( R_t \) is to maximize the L1 distance between the anchor and negative samples in the feature space, while minimizing the L1 distance between the anchor and positive samples. This can be formulated as:
\begin{equation}
R_t = \sum_{i=1}^n \omega_i \frac{\|V_i(J) - V_i(f(I, \theta))\|_1}{\|V_i(I) - V_i(f(I, \theta))\|_1}
\end{equation}
where \( f(\cdot, \theta) \) denotes the parameterized dehazing network, \( V_i(\cdot) \), \( i = 1, 2, \dots, n \), represents the \( i \)-th latent feature extracted from the pre-trained VGG-19 \cite{simonyan2014very} network, and \( \omega_i \) corresponds to their respective weights. However, as the anchor progressively shifts away from the negative samples and moves closer to the positive ones, the model becomes increasingly capable of distinguishing dehazed images from hazy inputs. Consequently, the influence of hazy images diminishes, limiting their effectiveness in further enhancing the model’s performance. To alleviate this issue and promote the generation of more natural dehazed outputs, we introduce a self-guided contrastive regularization strategy. Specifically, the coarse restored image \( \hat{I}_0 \) produced by the low-frequency restoration network, is treated as a hard negative sample. This design leverages the model's own intermediate output to encourage the final prediction to better approximate the ground truth. The regularization is formally defined as:
\begin{equation}
R_{SG} = \sum_{i=1}^n \omega_i \frac{\|V_i(J) - V_i(f(I, \theta))\|_1}{\|V_i(\hat{I}_0) - V_i(f(I, \theta))\|_1}
\end{equation}
\begin{table*}[t]
	\centering
	\caption{Quantitative comparisons of WDMamba and 12 SOTA dehazing methods on the Haze4K, RESIDE-6K, HSTS-Synthetic, and O-HAZE datasets. We report PSNR, SSIM, the number of parameters, and the number of FLOPs. Bold indicates the best results.}
	\label{tab:metrics}
	\begin{threeparttable}
		\footnotesize
		\centering
		\setlength{\tabcolsep}{1.5mm}{
			\begin{tabular}{cccccccccccc}
				\toprule
				\multirow{2}{*}{Method} &
                \multirow{2}{*}{Publication} &
				\multicolumn{2}{c}{Haze4K} & 
				\multicolumn{2}{c}{RESIDE-6K} & 
				\multicolumn{2}{c}{HSTS-Synthetic} & 
                \multicolumn{2}{c}{O-HAZE} &
				\multirow{2}{*}{Params (M)} & 
                \multirow{2}{*}{FLOPs (G)} \\ 
				\cmidrule(lr){3-4} \cmidrule(lr){5-6} \cmidrule(lr){7-8} \cmidrule(lr){9-10}
                & & PSNR$\uparrow$ & SSIM$\uparrow$ & PSNR$\uparrow$ & SSIM$\uparrow$ & PSNR$\uparrow$ & SSIM$\uparrow$ & PSNR$\uparrow$ & SSIM$\uparrow$ &  &  \\ 
				\midrule
				DCP \cite{he2010single} & TPAMI'10 & 16.93 & 0.5877 & 17.88 & 0.8160 & 17.01 & 0.8030 & 14.65 & 0.6358 & - & - \\ 
				AOD-Net \cite{li2017aod} & ICCV'17 & 17.90 & 0.5946 & 19.88 & 0.8454 & 20.87 & 0.8411 & 18.19 & 0.6823 & 0.002 & 0.12\\ 
				GDN \cite{liu2019griddehazenet} & ICCV'19 & 25.72 & 0.9641 & 27.16 & 0.9544 & 29.71 & 0.9617 & 20.05 & 0.7362 & 0.96 & 21.55 \\ 
				FFA-Net \cite{qin2020ffa} & AAAI'20 & 28.21 & 0.9669 & 28.69 & 0.9577 & 28.82 & 0.9133 & 23.34 & 0.8084 & 4.46 & 287.8 \\ 
				Dehamer \cite{guo2022dehamer} & CVPR'22 & 26.03 & 0.9392 & 28.12 & 0.9521 & 29.58 & 0.9207 & 24.36 & 0.8089 & 132.45 & 59.31\\ 
                DehazeFormer-L \cite{song2023vision} & TIP'23 & 31.86 & 0.9783 & 31.57 & 0.9696 & 34.08 & \textbf{0.9743} & 25.25 & 0.8206 & 25.45 & 277.02\\ 
				IR-SDE \cite{luo2023image} & ICML'23 & 29.57 & 0.9744 & 28.50 & 0.9575 & 27.60 & 0.8900 & 23.99 & 0.7652 & 135.3 & 119.1\\ 
                FSNet \cite{cui2023image} & TPAMI'23 & 34.09 & 0.9881 & 30.69 & 0.9672 & 30.54 & 0.9293 & 24.55 & 0.8483 & 13.28 & 110.5\\ 
				PNE-Net \cite{cheng2024progressive} & TMM'24 & 31.07 & 0.9821 & 29.64 & 0.9635 & 28.81 & 0.9523 & 24.12 & 0.8354 & 4.76 & 308.31\\ 
				DEA-Net \cite{chen2024dea} & TIP'24 & 34.22 & 0.9879 & 30.77 & 0.9681 & 31.66 & 0.9342 & 25.54 & 0.8196 & 3.65 & 32.23\\ 
				OKNet \cite{cui2024omni} & TCSVT'24 & 32.42 & 0.9863 & 30.21 & 0.9613 & 31.39 & 0.9316 & 25.62 & 0.8528 & 14.3 & 39.71\\ 
				ConvIR-B \cite{cui2024revitalizing} & TPAMI'24 & 34.12 & 0.9877 & 30.96 & 0.9656 & 31.80 & 0.9347 & 26.09 & 0.8552 & 8.63 & 71.22\\ 
                ConvIR-L \cite{cui2024revitalizing} & TPAMI'24 & 34.50 & 0.9886 & 30.23 & 0.9504 & 31.82 & 0.9330 & 25.31 & 0.8511 & 14.83 & 129.34 \\ 
				WDMamba  & Ours & \textbf{35.88} & \textbf{0.9909} & \textbf{32.15} & \textbf{0.9723} & \textbf{34.53} & 0.9739 & \textbf{27.22} & \textbf{0.8729} & 11.25 & 38.84 \\ 
				\bottomrule
			\end{tabular}
		}
	\end{threeparttable}
\end{table*}
\begin{figure*}[t] \centering
 	\includegraphics[width=1.0\linewidth]{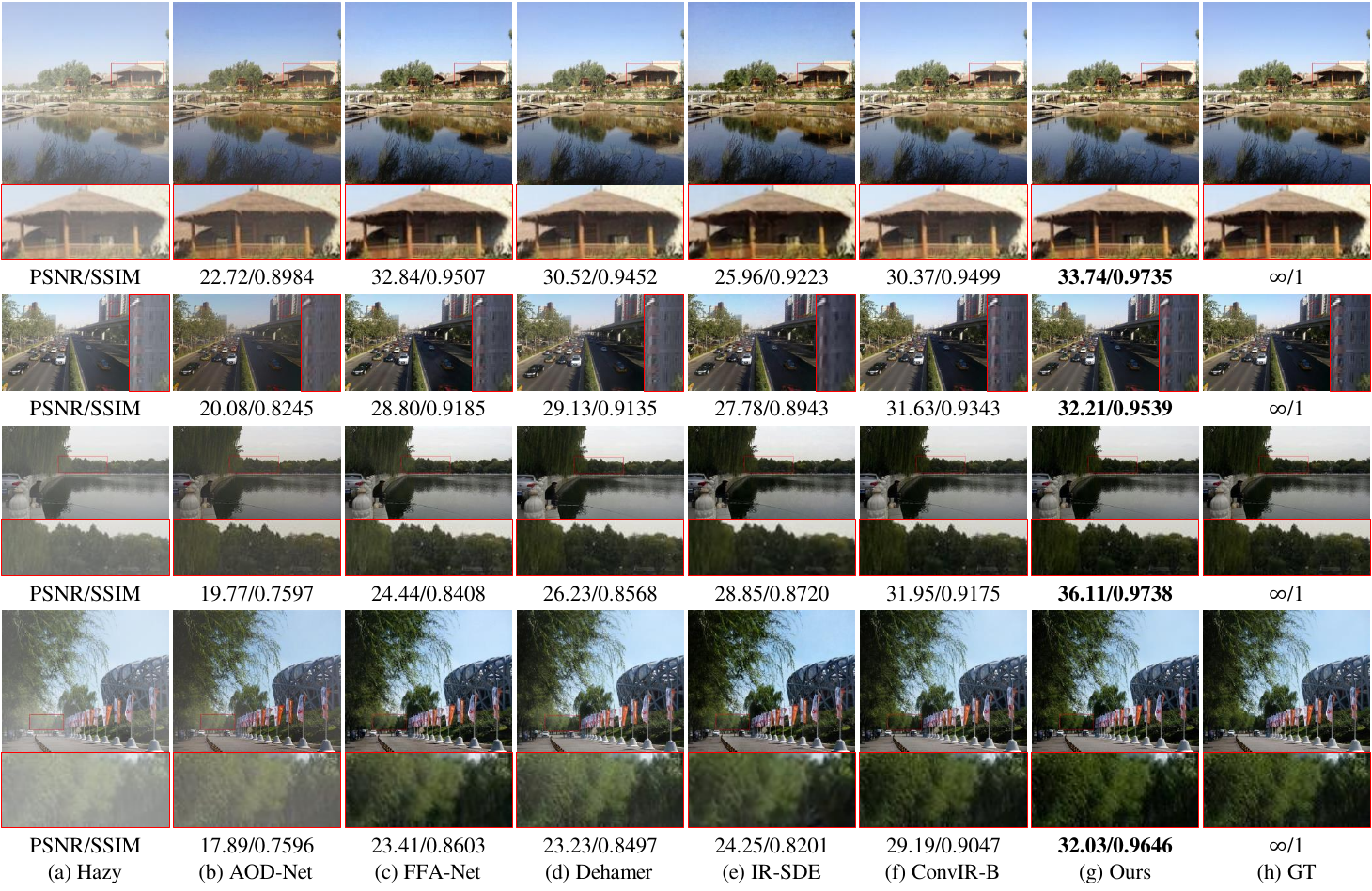}
 	\caption{ Visual comparisons on the RESIDE-6K and HSTS datasets. Our WDMamba demonstrates the ability to generate clearer images with richer details.}
 	\label{fig:syn_haze}
\end{figure*}
\begin{figure*}[t] \centering
 	\includegraphics[width=1.0\linewidth]{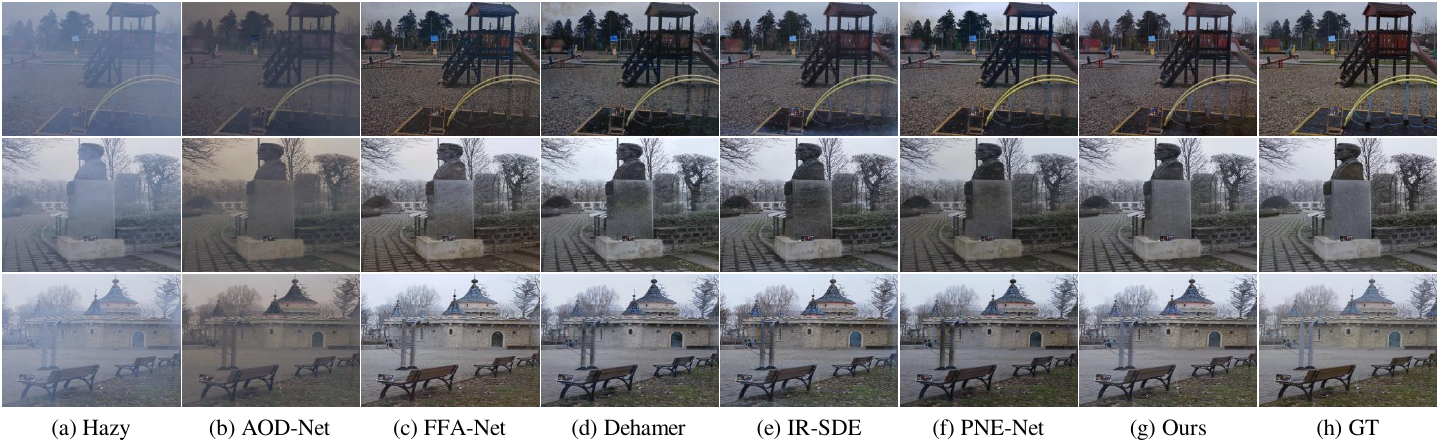}
 	\caption{ Visual comparisons on the O-HAZE dataset. Our WDMamba effectively removes haze while preserving superior color fidelity.}
 	\label{fig:ohaze}
\end{figure*}
\begin{figure*}[t] \centering
 	\includegraphics[width=1.0\linewidth]{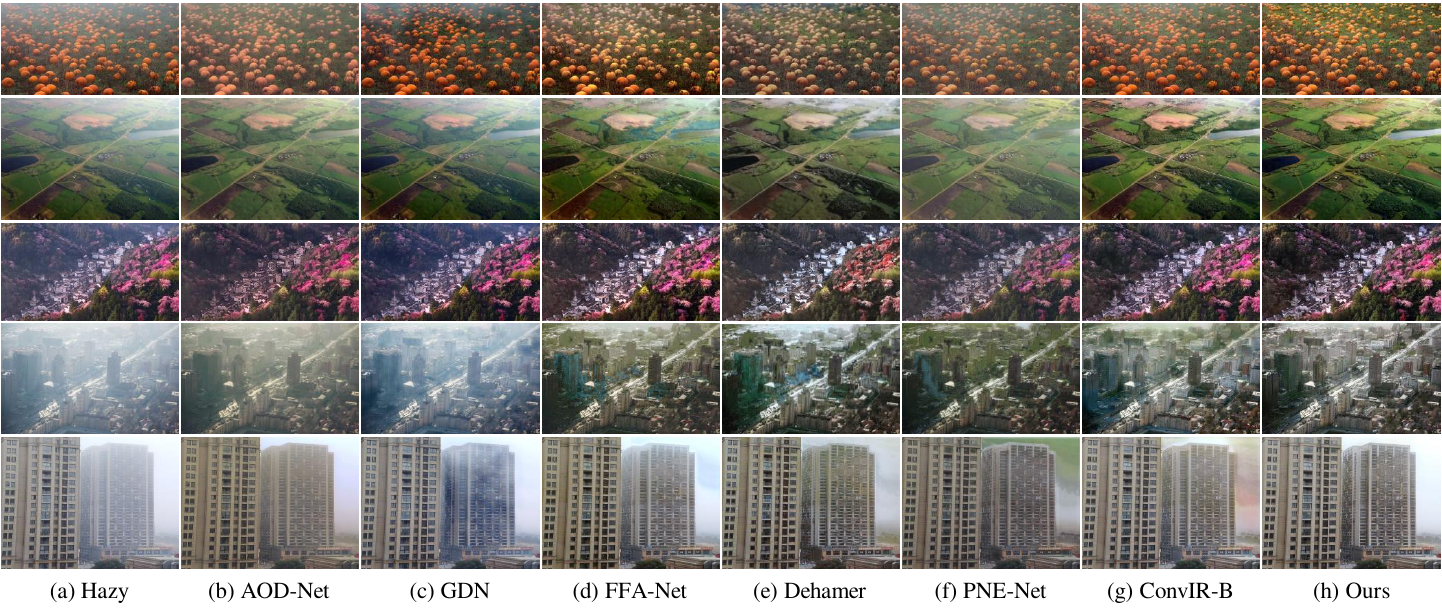}
 	\caption{ Visual comparisons on the real-world hazy images. The first two images are sourced from the Fattal's \cite{fattal2014dehazing} dataset, the third to four images are collected from the Internet, and the final image is captured using a mobile phone in hazy weather conditions. Our WDMamba is capable of producing clearer and more natural dehazed images.}
 	\label{fig:real_haze}
\end{figure*}
\subsection{Loss Function}
We adopt a comprehensive loss strategy that integrates spatial domain loss, frequency domain loss, and contrastive loss to effectively guide network training. In the spatial domain, we utilize L1 loss to supervise the final dehazed image \(Y\) with respect to the ground truth \( J \), defined as:
\begin{equation}
L_{spa} = {\|Y - J \|_1}
\end{equation}
For the frequency domain loss, we compute the L1 distance separately on the amplitude spectrum and phase spectrum of the dehazed image and the ground truth, expressed as:
\begin{equation}
L_{fre} = \| \mathcal{A}(Y) - \mathcal{A}(J) \|_1 + \| \mathcal{P}(Y) - \mathcal{P}(J) \|_1
\end{equation}
Furthermore, the contrastive loss includes both the conventional contrastive regularization and the proposed self-guided contrastive regularization, formulated as:
\begin{equation}
L_{cr} = (1-\mu)R_t + (1+\mu)R_{SG}
\end{equation}
where \(\mu\) is a balancing factor empirically set to 0.25. Finally, the total loss function is defined as a weighted combination of the above components:
\begin{equation}
L_{total} = L_{spa} + \lambda_1L_{fre} + \lambda_2L_{cr}
\end{equation}
where \(\lambda_1\) and \(\lambda_2\) are empirically set to 0.1 and 0.001.

\section{EXPERIMENTS}
In this section, we first introduce the experimental setup, followed by extensive experiments to demonstrate the effectiveness of the proposed method. Finally, we conduct a detailed ablation analysis of the proposed approach.
\subsection{Experiment Setting}
\subsubsection{Datasets}
To facilitate a comprehensive comparison with existing dehazing methods, both quantitative and qualitative evaluations are conducted using three synthetic datasets: Haze4K \cite{liu2021synthetic}, RESIDE-6K \cite{shao2020domain}, and HSTS (Hybrid Subjective Testing Set), along with a real-world hazy dataset, O-HAZE \cite{ancuti2018haze}. Haze4K comprises 3,000 training images and 1,000 test images, whereas RESIDE-6K includes 6,000 training images and 1,000 test images. Both datasets contain test sets featuring a mix of indoor and outdoor hazy images. The HSTS synthetic dataset consists of 10 test images without any training data. In this study, evaluation on the HSTS dataset is performed using the model trained on RESIDE-6K. O-HAZE provides 45 real-world outdoor hazy images, which are manually generated, with the first 40 allocated for training and the remaining 5 for testing. All O-HAZE images are resized to a resolution of \(1600\times1200\) pixels for evaluation.

\subsubsection{Implementation \ details}
Our model was implemented using the PyTorch framework on an NVIDIA RTX 3090 GPU. We set the number of Mamba blocks in each layer to \([1, 1, 2, 2, 4, 4, 2, 2, 1]\). For synthetic hazy datasets, the number of block structures \( N \) in detail enhancement network is set to 6. In real image dehazing experiments, \( N \) is set to 4 due to the smaller number of training images. We employ a progressive training strategy. Taking the RESIDE-6K dataset as an example, we set the total number of iterations to 500,000, with image sizes of \([256, 400]\) and corresponding batch sizes of \([12, 4]\). The initial learning rate is set to \( 5 \times 10^{-4} \) and gradually decays to \( 1 \times 10^{-7} \) using the cosine annealing strategy.

\subsubsection{Comparison Methods and Evaluation Metrics}
We compare our WDMamba with two earlier methods, including DCP \cite{he2010single} and AOD-Net \cite{li2017aod}, as well as ten recent competing methods, including GDN \cite{liu2019griddehazenet}, FFA-Net \cite{qin2020ffa}, Dehamer \cite{guo2022dehamer}, DehazeFormer \cite{song2023vision}, IR-SDE \cite{luo2023image}, FSNet \cite{cui2023image}, PNE-Net \cite{cheng2024progressive}, DEA-Net \cite{chen2024dea}, OKNet \cite{cui2024omni}, and ConvIR \cite{cui2024revitalizing}. Furthermore, we use Peak Signal to Noise Ratio (PSNR) and  Structural Similarity Index (SSIM) \cite{assessment2004error} to evaluate dehazing performance.
\subsection{Comparison with State-of-the-art Methods}
\subsubsection{Evaluation on Synthetic Hazy Images} Table \ref{tab:metrics} presents the quantitative comparison between our approach and other SOTA methods across three synthetic datasets. As observed, our WDMamba achieving a 1.38 dB improvement over ConvIR-L on the Haze4K dataset and a 0.58 dB gain over DehazeFormer-L on the RESIDE-6K dataset. Moreover, WDMamba also demonstrates strong competitiveness on the HSTS-synthetic dataset, achieving the best PSNR and a SSIM score comparable to DehazeFormer-L. These extensive quantitative comparisons confirm that WDMamba consistently delivers improved dehazing performance. We also present a visual comparison of dehazed images produced by representative algorithms in Fig. \ref{fig:syn_haze}, from which it can be observed that our method more effectively restores fine details and produces clearer, sharper outputs. In particular, it exhibits superior dehazing capability in distant regions, where haze is typically more challenging to remove.
\subsubsection{Evaluation on Real Hazy Images} We further evaluate the proposed WDMamba against SOTA methods on a real-world hazy dataset. The quantitative comparison is presented in Table \ref{tab:metrics}, where our method achieves a performance gain of 1.13 dB over ConvIR-B. A visual comparison is provided in Fig. \ref{fig:ohaze}. As can be seen, for the real-world hazy dataset characterized by higher haze density, the traditional method AOD-Net exhibits poor dehazing performance. Although more advanced deep learning-based approaches are able to remove most of the haze, their results often suffer from noticeable color distortions in certain regions. In contrast, our approach not only effectively eliminates haze but also achieves better restoration of local details, resulting in dehazed images with enhanced color fidelity. Beyond the benchmark dataset, we further evaluate WDMamba on real-world hazy images without reference ground truth under diverse haze scenarios. Fig. \ref{fig:real_haze} presents 5 real-world hazy images along with the corresponding dehazing results produced by various methods. As shown, WDMamba generates clearer and more visually faithful dehazing results, demonstrating improved adaptability.

\subsection{Ablation Study}
\begin{table}[t]
    \centering
    \footnotesize % 缩小表格字体
    \begin{threeparttable} % 使用 threeparttable 结构
        \caption{Ablation studies of coarse to fine architecture.}
        \label{tab:ablation_two_stage}
        \begin{tabular}{lccccccc}
            \toprule
            Model & LFRN & DEN & PSNR & SSIM & Params (M) & FLOPs (G)\\
            \midrule
           \(M_1\) & \checkmark & & 29.56 & 0.9573 & 5.53 & 1.28 \\
           \(M_2\) & & \checkmark & 29.75 & 0.9644 & 5.72 & 37.56 \\
            \(M_3\) & RDB & & 29.18 & 0.9566 & 3.74 & 3.36 \\
            \(M_4\) & RDB & \checkmark & 30.68 & 0.9691 & 9.46 & 40.93 \\
            \(M_5\) & \checkmark & \checkmark & \textbf{32.15} & \textbf{0.9723} & 11.25 & 38.84 \\
            \bottomrule
        \end{tabular}
    \end{threeparttable}
\end{table}
In this section, we conduct an ablation study on our key design using the RESIDE-6K dataset.
\subsubsection{Coarse to Fine Architecture} We first conduct an ablation study to evaluate the effectiveness of the low-frequency restoration network and the detail enhancement network within our coarse-to-fine architecture. The experimental results are presented in Table \ref{tab:ablation_two_stage}. Models \(M_1\) and \(M_2\) employ only the low-frequency restoration network and the detail enhancement network, respectively. The results indicate that a single-stage network is insufficient for high-quality image reconstruction. Furthermore, we conduct a comparative experiment by replacing the Mamba blocks in LFRN with RDBs \cite{zhang2018residual} (models \(M_3\) and \(M_4\)), which have comparable numbers of parameters and FLOPs. The results demonstrate that, compared to RDB, the Mamba block leverages the strength of state space models in capturing long-range spatial dependencies, thereby achieving superior performance.

\subsubsection{Self-Guided Contrastive Regularization} We further conduct an ablation study on contrastive regularization, with the experimental results presented in Table \ref{tab:SGCR}. It can be observed that the conventional contrastive paradigm provides limited performance gains. In comparison, the proposed self-guided contrastive regularization enhances the model’s feature discriminability by treating the coarse restored image as informative guidance, thereby leading to improved dehazing results.
\begin{table}[t]
    \centering
    \begin{threeparttable}
        \caption{Ablation study of the contribution of contrastive regularization}
        \label{tab:SGCR}
        \begin{tabular}{lcccc}
            \toprule
            & w/o CR & CR & SGCR & CR + SGCR \\
            \midrule
            PSNR & 31.46 & 31.60 & 31.91 & 32.15 \\
            SSIM & 0.9711 & 0.9716 & 0.9722 & 0.9723 \\
            \bottomrule
        \end{tabular}
    \end{threeparttable}
\end{table}

\section{CONCLUSION}
In this work, we propose WDMamba, a coarse-to-fine image dehazing framework inspired by the prior observation that haze degradation predominantly resides in the low-frequency components of the wavelet transform. Guided by this prior knowledge, WDMamba effectively learns task-adaptive feature representations across distinct stages, thereby enabling highly effective image dehazing. Specifically, the framework first employs a Mamba-based Low-Frequency Restoration Network to reconstruct global structures, followed by a CNN-based Detail Enhancement Network to recover fine-grained details. This architecture effectively exploits the capability of Vision State Space Blocks to model long-range spatial dependencies with linear complexity, while leveraging the local detail extraction strengths of CNNs. In addition, we introduce a self-guided contrastive regularization strategy during training, which further boosts the model's dehazing efficacy. Extensive experimental results demonstrate that our method consistently outperforms existing state-of-the-art approaches across multiple public benchmarks.

\vspace{5pt}


\bibliographystyle{IEEEtran}
\bibliography{reference}

\begin{thebibliography}{10}
\providecommand{\url}[1]{#1}
\csname url@samestyle\endcsname
\providecommand{\newblock}{\relax}
\providecommand{\bibinfo}[2]{#2}
\providecommand{\BIBentrySTDinterwordspacing}{\spaceskip=0pt\relax}
\providecommand{\BIBentryALTinterwordstretchfactor}{4}
\providecommand{\BIBentryALTinterwordspacing}{\spaceskip=\fontdimen2\font plus
\BIBentryALTinterwordstretchfactor\fontdimen3\font minus \fontdimen4\font\relax}
\providecommand{\BIBforeignlanguage}[2]{{%
\expandafter\ifx\csname l@#1\endcsname\relax
\typeout{** WARNING: IEEEtran.bst: No hyphenation pattern has been}%
\typeout{** loaded for the language `#1'. Using the pattern for}%
\typeout{** the default language instead.}%
\else
\language=\csname l@#1\endcsname
\fi
#2}}
\providecommand{\BIBdecl}{\relax}
\BIBdecl

\bibitem{liu2018improved}
Y.~Liu, G.~Zhao, B.~Gong, Y.~Li, R.~Raj, N.~Goel, S.~Kesav, S.~Gottimukkala, Z.~Wang, W.~Ren \emph{et~al.}, ``Improved techniques for learning to dehaze and beyond: A collective study,'' \emph{arXiv preprint arXiv:1807.00202}, 2018.

\bibitem{ren2018deep}
W.~Ren, J.~Zhang, X.~Xu, L.~Ma, X.~Cao, G.~Meng, and W.~Liu, ``Deep video dehazing with semantic segmentation,'' \emph{{IEEE} Trans. Image Process.}, vol.~28, no.~4, pp. 1895--1908, 2018.

\bibitem{he2010single}
K.~He, J.~Sun, and X.~Tang, ``Single image haze removal using dark channel prior,'' \emph{{IEEE} Trans. Pattern Anal. Mach. Intell.}, vol.~33, no.~12, pp. 2341--2353, Sep. 2010.

\bibitem{meng2013efficient}
G.~Meng, Y.~Wang, J.~Duan, S.~Xiang, and C.~Pan, ``Efficient image dehazing with boundary constraint and contextual regularization,'' in \emph{Proc. IEEE Int. Conf. Comput. Vis. (ICCV)}, 2013, pp. 617--624.

\bibitem{zhu2015fast}
Q.~Zhu, J.~Mai, and L.~Shao, ``A fast single image haze removal algorithm using color attenuation prior,'' \emph{{IEEE} Trans. Image Process.}, vol.~24, no.~11, pp. 3522--3533, 2015.

\bibitem{berman2016non}
D.~Berman, S.~Avidan \emph{et~al.}, ``Non-local image dehazing,'' in \emph{Proc. IEEE/CVF Conf. Comput. Vis. Pattern Recognit. (CVPR)}, 2016, pp. 1674--1682.

\bibitem{ju2021idbp}
M.~Ju, C.~Ding, W.~Ren, and Y.~Yang, ``Idbp: Image dehazing using blended priors including non-local, local, and global priors,'' \emph{{IEEE} Trans. Circuits Syst. Video Technol.}, vol.~32, no.~7, pp. 4867--4871, 2021.

\bibitem{narasimhan2000chromatic}
S.~G. Narasimhan and S.~K. Nayar, ``Chromatic framework for vision in bad weather,'' in \emph{Proc. IEEE/CVF Conf. Comput. Vis. Pattern Recognit. (CVPR)}, vol.~1, 2000, pp. 598--605.

\bibitem{liu2024vmamba}
Y.~Liu, Y.~Tian, Y.~Zhao, H.~Yu, L.~Xie, Y.~Wang, Q.~Ye, J.~Jiao, and Y.~Liu, ``Vmamba: Visual state space model,'' \emph{Proc. Annu. Conf. Neural Inf. Process. Syst. (NeurIPS)}, vol.~37, pp. 103\,031--103\,063, 2024.

\bibitem{ma2024u}
J.~Ma, F.~Li, and B.~Wang, ``U-mamba: Enhancing long-range dependency for biomedical image segmentation,'' \emph{arXiv preprint arXiv:2401.04722}, 2024.

\bibitem{zou2024freqmamba}
Z.~Zou, H.~Yu, J.~Huang, and F.~Zhao, ``Freqmamba: Viewing mamba from a frequency perspective for image deraining,'' in \emph{Proc. ACM Int. Conf. Multimedia (ACM MM)}, 2024, pp. 1905--1914.

\bibitem{liu2021synthetic}
Y.~Liu, L.~Zhu, S.~Pei, H.~Fu, J.~Qin, Q.~Zhang, L.~Wan, and W.~Feng, ``From synthetic to real: Image dehazing collaborating with unlabeled real data,'' in \emph{Proc. ACM Int. Conf. Multimedia (ACM MM)}, 2021, pp. 50--58.

\bibitem{liu2019griddehazenet}
X.~Liu, Y.~Ma, Z.~Shi, and J.~Chen, ``Griddehazenet: Attention-based multi-scale network for image dehazing,'' in \emph{Proc. IEEE Int. Conf. Comput. Vis. (ICCV)}, 2019, pp. 7314--7323.

\bibitem{dong2020multi}
H.~Dong, J.~Pan, L.~Xiang, Z.~Hu, X.~Zhang, F.~Wang, and M.-H. Yang, ``Multi-scale boosted dehazing network with dense feature fusion,'' in \emph{Proc. IEEE/CVF Conf. Comput. Vis. Pattern Recognit. (CVPR)}, 2020, pp. 2157--2167.

\bibitem{qin2020ffa}
X.~Qin, Z.~Wang, Y.~Bai, X.~Xie, and H.~Jia, ``Ffa-net: Feature fusion attention network for single image dehazing,'' in \emph{Proc. AAAI Conf. Artif. Intell. (AAAI)}, vol.~34, no.~07, 2020, pp. 11\,908--11\,915.

\bibitem{zhang2021hierarchical}
X.~Zhang, J.~Wang, T.~Wang, and R.~Jiang, ``Hierarchical feature fusion with mixed convolution attention for single image dehazing,'' \emph{{IEEE} Trans. Circuits Syst. Video Technol.}, vol.~32, no.~2, pp. 510--522, 2021.

\bibitem{zhang2023sdbad}
G.~Zhang, W.~Fang, Y.~Zheng, and R.~Wang, ``Sdbad-net: A spatial dual-branch attention dehazing network based on meta-former paradigm,'' \emph{{IEEE} Trans. Circuits Syst. Video Technol.}, vol.~34, no.~1, pp. 60--70, 2023.

\bibitem{song2023vision}
Y.~Song, Z.~He, H.~Qian, and X.~Du, ``Vision transformers for single image dehazing,'' \emph{{IEEE} Trans. Image Process.}, vol.~32, pp. 1927--1941, 2023.

\bibitem{liu2021swin}
Z.~Liu, Y.~Lin, Y.~Cao, H.~Hu, Y.~Wei, Z.~Zhang, S.~Lin, and B.~Guo, ``Swin transformer: Hierarchical vision transformer using shifted windows,'' in \emph{Proc. IEEE Int. Conf. Comput. Vis. (ICCV)}, 2021, pp. 10\,012--10\,022.

\bibitem{wang2023uscformer}
Y.~Wang, J.~Xiong, X.~Yan, and M.~Wei, ``Uscformer: Unified transformer with semantically contrastive learning for image dehazing,'' \emph{{IEEE} Trans. Intell. Transp. Syst.}, vol.~24, no.~10, pp. 11\,321--11\,333, 2023.

\bibitem{gu2021efficiently}
A.~Gu, K.~Goel, and C.~R{\'e}, ``Efficiently modeling long sequences with structured state spaces,'' \emph{arXiv preprint arXiv:2111.00396}, 2021.

\bibitem{smith2022simplified}
J.~T. Smith, A.~Warrington, and S.~W. Linderman, ``Simplified state space layers for sequence modeling,'' \emph{arXiv preprint arXiv:2208.04933}, 2022.

\bibitem{gu2023mamba}
A.~Gu and T.~Dao, ``Mamba: Linear-time sequence modeling with selective state spaces,'' \emph{arXiv preprint arXiv:2312.00752}, 2023.

\bibitem{zou2024wave}
W.~Zou, H.~Gao, W.~Yang, and T.~Liu, ``Wave-mamba: Wavelet state space model for ultra-high-definition low-light image enhancement,'' in \emph{Proc. ACM Int. Conf. Multimedia (ACM MM)}, 2024, pp. 1534--1543.

\bibitem{oord2018representation}
A.~v.~d. Oord, Y.~Li, and O.~Vinyals, ``Representation learning with contrastive predictive coding,'' \emph{arXiv preprint arXiv:1807.03748}, 2018.

\bibitem{sermanet2018time}
P.~Sermanet, C.~Lynch, Y.~Chebotar, J.~Hsu, E.~Jang, S.~Schaal, S.~Levine, and G.~Brain, ``Time-contrastive networks: Self-supervised learning from video,'' in \emph{Proc. IEEE Int. Conf. Robot. Autom. (ICRA)}, 2018, pp. 1134--1141.

\bibitem{chen2020simple}
T.~Chen, S.~Kornblith, M.~Norouzi, and G.~Hinton, ``A simple framework for contrastive learning of visual representations,'' in \emph{Proc. Int. Conf. Mach. Learn. (ICML)}, 2020, pp. 1597--1607.

\bibitem{he2020momentum}
K.~He, H.~Fan, Y.~Wu, S.~Xie, and R.~Girshick, ``Momentum contrast for unsupervised visual representation learning,'' in \emph{Proc. IEEE/CVF Conf. Comput. Vis. Pattern Recognit. (CVPR)}, 2020, pp. 9729--9738.

\bibitem{zheng2023curricular}
Y.~Zheng, J.~Zhan, S.~He, J.~Dong, and Y.~Du, ``Curricular contrastive regularization for physics-aware single image dehazing,'' in \emph{Proc. IEEE/CVF Conf. Comput. Vis. Pattern Recognit. (CVPR)}, 2023, pp. 5785--5794.

\bibitem{feng2024advancing}
Y.~Feng, L.~Ma, X.~Meng, F.~Zhou, R.~Liu, and Z.~Su, ``Advancing real-world image dehazing: perspective, modules, and training,'' \emph{{IEEE} Trans. Pattern Anal. Mach. Intell.}, 2024.

\bibitem{gao2024efficient}
N.~Gao, X.~Jiang, X.~Zhang, and Y.~Deng, ``Efficient frequency-domain image deraining with contrastive regularization,'' in \emph{Proc. Eur. Conf. Comput. Vis. (ECCV)}, 2024, pp. 240--257.

\bibitem{wang2024ucl}
Y.~Wang, X.~Yan, F.~L. Wang, H.~Xie, W.~Yang, X.-P. Zhang, J.~Qin, and M.~Wei, ``Ucl-dehaze: toward real-world image dehazing via unsupervised contrastive learning,'' \emph{{IEEE} Trans. Image Process.}, vol.~33, pp. 1361--1374, 2024.

\bibitem{ronneberger2015u}
O.~Ronneberger, P.~Fischer, and T.~Brox, ``U-net: Convolutional networks for biomedical image segmentation,'' in \emph{Proc. Int. Conf. Med. Image Comput. Comput.-Assist. Intervent (MICCAI)}, 2015, pp. 234--241.

\bibitem{xu2023haar}
G.~Xu, W.~Liao, X.~Zhang, C.~Li, X.~He, and X.~Wu, ``Haar wavelet downsampling: A simple but effective downsampling module for semantic segmentation,'' \emph{Pattern Recognition}, vol. 143, p. 109819, 2023.

\bibitem{wang2020eca}
Q.~Wang, B.~Wu, P.~Zhu, P.~Li, W.~Zuo, and Q.~Hu, ``Eca-net: Efficient channel attention for deep convolutional neural networks,'' in \emph{Proc. IEEE/CVF Conf. Comput. Vis. Pattern Recognit. (CVPR)}, 2020, pp. 11\,534--11\,542.

\bibitem{he2016deep}
K.~He, X.~Zhang, S.~Ren, and J.~Sun, ``Deep residual learning for image recognition,'' in \emph{Proc. IEEE/CVF Conf. Comput. Vis. Pattern Recognit. (CVPR)}, 2016, pp. 770--778.

\bibitem{simonyan2014very}
K.~Simonyan, ``Very deep convolutional networks for large-scale image recognition,'' \emph{arXiv preprint arXiv:1409.1556}, 2014.

\bibitem{li2017aod}
B.~Li, X.~Peng, Z.~Wang, J.~Xu, and D.~Feng, ``Aod-net: All-in-one dehazing network,'' in \emph{Proc. IEEE Int. Conf. Comput. Vis. (ICCV)}, 2017, pp. 4770--4778.

\bibitem{guo2022dehamer}
C.-L. Guo, Q.~Yan, S.~Anwar, R.~Cong, W.~Ren, and C.~Li, ``Image dehazing transformer with transmission-aware 3d position embedding,'' in \emph{Proc. IEEE/CVF Conf. Comput. Vis. Pattern Recognit. (CVPR)}, 2022, pp. 5812--5820.

\bibitem{luo2023image}
Z.~Luo, F.~K. Gustafsson, Z.~Zhao, J.~Sj{\"o}lund, and T.~B. Sch{\"o}n, ``Image restoration with mean-reverting stochastic differential equations,'' \emph{Proc. Int. Conf. Mach. Learn. (ICML)}, 2023.

\bibitem{cui2023image}
Y.~Cui, W.~Ren, X.~Cao, and A.~Knoll, ``Image restoration via frequency selection,'' \emph{{IEEE} Trans. Pattern Anal. Mach. Intell.}, 2023.

\bibitem{cheng2024progressive}
D.~Cheng, Y.~Li, D.~Zhang, N.~Wang, J.~Sun, and X.~Gao, ``Progressive negative enhancing contrastive learning for image dehazing and beyond,'' \emph{IEEE Trans. Multimedia}, 2024.

\bibitem{chen2024dea}
Z.~Chen, Z.~He, and Z.-M. Lu, ``Dea-net: Single image dehazing based on detail-enhanced convolution and content-guided attention,'' \emph{{IEEE} Trans. Image Process.}, 2024.

\bibitem{cui2024omni}
Y.~Cui, W.~Ren, and A.~Knoll, ``Omni-kernel modulation for universal image restoration,'' \emph{{IEEE} Trans. Circuits Syst. Video Technol.}, 2024.

\bibitem{cui2024revitalizing}
Y.~Cui, W.~Ren, X.~Cao, and A.~Knoll, ``Revitalizing convolutional network for image restoration,'' \emph{{IEEE} Trans. Pattern Anal. Mach. Intell.}, 2024.

\bibitem{fattal2014dehazing}
R.~Fattal, ``Dehazing using color-lines,'' \emph{{ACM} Trans. Graph.}, vol.~34, no.~1, pp. 1--14, 2014.

\bibitem{shao2020domain}
Y.~Shao, L.~Li, W.~Ren, C.~Gao, and N.~Sang, ``Domain adaptation for image dehazing,'' in \emph{Proc. IEEE/CVF Conf. Comput. Vis. Pattern Recognit. (CVPR)}, 2020, pp. 2808--2817.

\bibitem{ancuti2018haze}
C.~O. Ancuti, C.~Ancuti, R.~Timofte, and C.~De~Vleeschouwer, ``O-haze: a dehazing benchmark with real hazy and haze-free outdoor images,'' in \emph{Proc. IEEE/CVF Conf. Comput. Vis. Pattern Recognit. workshops (CVPRW)}, 2018, pp. 754--762.

\bibitem{assessment2004error}
I.~Q. Assessment, ``From error visibility to structural similarity,'' \emph{{IEEE} Trans. Image Process.}, vol.~13, no.~4, p.~93, 2004.

\bibitem{zhang2018residual}
Y.~Zhang, Y.~Tian, Y.~Kong, B.~Zhong, and Y.~Fu, ``Residual dense network for image super-resolution,'' in \emph{Proc. IEEE/CVF Conf. Comput. Vis. Pattern Recognit. (CVPR)}, 2018, pp. 2472--2481.

\end{thebibliography}

\vfill

\end{document}